# Investigating the Impact of Personalized AI Tutors on Language Learning Performance

Simon Suh
Department of Technology and Society
Stony Brook University

**[Abstract]**

Driven by the global shift towards online learning prompted by the COVID-19 pandemic, Artificial Intelligence (AI) has emerged as a pivotal player in the field of education. Intelligent Tutoring Systems (ITS) offer a new method of personalized teaching, replacing the limitations of traditional teaching methods. However, concerns arise about the ability of AI tutors to address skill development and engagement during the learning process. In this paper, I will conduct a quasi-experiment with paired-sample t-test on 34 students pre- and post-use of AI tutors in language learning platforms like Santa and Duolingo to examine the relationship between students' engagement, academic performance, and students' satisfaction during a personalized language learning experience.

# 1. Introduction

The educational landscape is undergoing a transformative shift with the integration of Artificial Intelligence (AI). Technologies like Intelligent Tutoring Systems (ITS), specifically designed to provide individualized instruction and feedback to learners (Sedlmeier, 2002), play a crucial role in this transformation, steering the educational landscape towards the Application of Artificial Intelligence in Education (AIEd) (Thomas et al., 2023). As the accessibility and diversity of AI technologies increase, this holds a significant potential to personalize learning experiences and unlock the educational potential of each student by fostering a more efficient and effective learning process (Rane, 2023).

Traditional teaching methods faced challenges in meeting the diverse needs and learning styles of individual students due to their dependence on rote memorization and a standardized, one-size-fits-all approach (Ballan, 2023; Lopez, 2008). Recognizing this limitation, the exploration of personalized learning approaches began, even before the adoption of AI technology. The Zone of Proximal Development (ZPD), introduced by psychologist and educational theorist Lev Vygotsky (1930), emphasized that a range of tasks that a learner can't do independently can be achieved with the appropriate support and guidance from a teacher or peer. Unlike traditional teaching methods which lack the fundamental elements for individual growth, ZPD highlights the importance of a more personalized educational approach. Personalized learning strives to address the specific needs of each student (Tetzlaff, 2020). This is where AI-powered personalized learning becomes crucial, aiming to provide unique learning pathways that maximize individual strengths and address specific weaknesses.

While existing research has laid the groundwork for the benefits of AI tutors in personalized education (Kim et al., 2020; Al Yakin Muthmainnah Ahmad et al., 2023; Thomas et al., 2023; Woo et al., 2021; Yildiz., 2023; Ruiz Rey et al., 2023), a study by Mollick (2023) raised concerns about biased, instructional, and confabulation risks of AI that may potentially limit skill development and engagement of learners. This study aims to bridge this gap by providing detailed insights into the impact of personalized AI tutors specifically in language learning and their potential application across a wider range of educational settings. Specifically, we will address the research question: how does a personalized language learning experience facilitated by AI tutors in language learning platforms like Santa and Duolingo influence students' engagement, academic performance, and students' satisfaction with the learning material? This study aims to provide empirical evidence on the effectiveness of AI tutors in diverse language learning contexts but also reveal the potential for integration across diverse educational settings.

## 2. Literature Review

### 2.1. Theoretical Framework

This research is grounded in three theoretical frameworks: Domain modeling (Minn, 2022), Gardner's Theory of Multiple Intelligences (GTMI) (Morgan, 2021), and Zone of Proximal Development (ZPD) (Vygotsky, 1930). Domain modeling serves as a critical component of Intelligent Tutoring Systems (ITS) in AI. The model extracts specific knowledge components within a particular subject area and uses them to build a student model that tracks the learner's strengths and weaknesses in domain-specific knowledge. This model enables ITS, like AI tutors, to personalize instruction, provide specific interventions, and offer individual feedback to enhance skill development (Minn, 2022). GTMI is a theory developed by developmental psychologist Howard Earl Gardner that provides the theoretical foundation for modern personalized learning (Morgan, 2021). The theory proposes that humans have eight distinct intelligences, rather than just one general intelligence. Each intelligence represents a different way of understanding and processing information, and individuals can excel in one or more of these areas (Davis, 2011). According to Morgan (2021), GTMI and ZPD are the foundational elements of personalized learning. It is crucial for instructors, whether AI or human, to understand these two theoretical frameworks to design and deliver personalized learning for learners. In the context of our research, we utilized these three theoretical frameworks to examine how AI tutors can facilitate a personalized language learning experience, aiming to influence outcomes related to students' engagement, academic performance, and students' satisfaction.

### 2.2. AI Tutors and Students' Engagement

In response to the global shift towards online learning prompted by the COVID-19 pandemic, Intelligent Tutoring Systems (ITS) have emerged as a promising solution to address the challenges of this transformed educational landscape. To support this, Wang et al. (2021) found that ITS has been applied to support learning in multiple subjects (see Figure 1) and 62.5% of studies conducted regarding the overall impact of ITS on learning have reported positive.

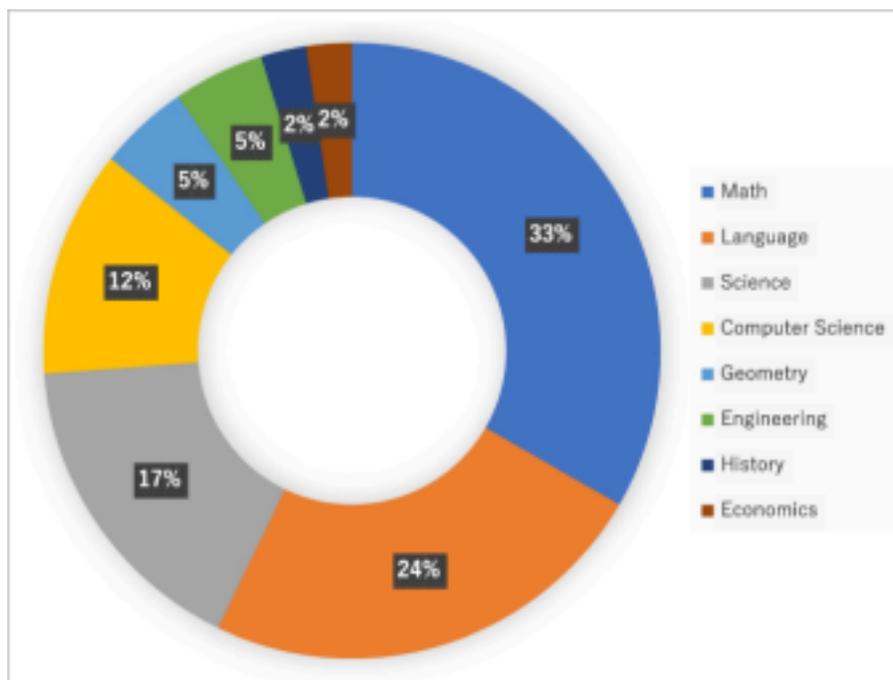

Figure 1: Distribution of subjects where ITS were found to support learning (Wang et al., 2021, pg.9128)

Recognizing the pivotal role of student engagement in online learning, Kim et al. (2020) investigated AI-driven ITS interface design, specifically focusing on diagnostic feedback for students' problem-solving processes. Using Santa, an AI tutoring service for TOEIC preparation in South Korea, they examined that student engagement increased by 25.13% when an AI provided specific analysis of the student's skills and recommendations for a personalized curriculum. Despite acknowledging limitations such as a focus on TOEIC preparation and sample size, the study showcased the potential of AI in enhancing engagement. Furthermore, Xu et al. (2022) revealed a high engagement level (4.5 out of 5) among young learners engaged in English language learning with AI tutors. The research utilized AI-measured indicators, specifically incorporating facial feedback, to evaluate and analyze how to improve individual learner engagement.

Building on this statement, Diwan et al. (2022) highlighted AI's potential to personalize learning pathways based on narrative fragment generation and natural language generation (NLG) -based question creation. The research proposed that these methods aim to make learning pathways more interactive, adaptable across domains, and dynamically respond to learner-driven changes. However, as the study acknowledged, realizing the full potential of AI requires careful consideration of accessibility for all students to navigate this new educational landscape.

Additionally, Ruan, S et al. (2021) investigated an AI-powered conversational chatbot called EnglishBot, which aimed to improve Chinese-speaking students' English-speaking skills. Their study found that the conversational aspect of EnglishBot allowed users to spend significantly more time engaged compared to a traditional listen-and-repeat interface. Moreover, EnglishBot users demonstrated significant improvements in vocabulary acquisition and speaking proficiency, showcasing the potential of AI tutors to effectively enhance language learning.

Overall, this review highlights AI's potential to transform user engagement and personalized education. From ITS problem-solving to chatbots enhanced language learning, these studies offer valuable insights for educators and researchers. While current research highlights the potential of AI in personalized language learning, critical research gaps remain in our understanding of its broader applicability and long-term impact. Existing studies often focus on specific language skills or educational settings, limiting the generalizability of findings. Furthermore, concerns regarding accessibility for diverse learners and the need for effective teacher training to support AI integration remain largely unaddressed. This research aims to fill these critical gaps by conducting a study that investigates the impact of personalized AI tutors across language learning contexts and educational settings.

## 2.3. AI Tutors and Academic Performance

Current research suggests a growing trend of AI integration in language learning tools, promising significant benefits for student performance. This analysis of two prominent studies reveals the specific mechanisms and outcomes associated with AI-powered language learning. Thomas et al. (2023) researched the impact of Artificial Intelligence (AI) on education. The research adhered to the Preferred Reporting Items for Systematic Reviews and Meta-Analyses (PRISMA) statement and the outcomes of the study identified four distinct learning enhancements facilitated by AI. The results showed that AI learning tools accustomed to students' individual needs not only foster their engagement but also enhance performance in language learning (Thomas et al., 2023)

Furthermore, a study by Woo et al. (2021) used PRISA guidelines to explore current trends in AI-based language learning tools. Their findings revealed the development of AI tools targeting various language skills had a connection to the improvement of learners' language abilities. A study that used a Kaizen language learning platform that uses a personalized AI tutor for conversation practice showed an increase in fluency and accuracy in both practice and test settings (Nakayama, 2022). Another study investigated how a personalized AI-assisted TOEIC learning program, such as Santa TOEIC and Soljam, helped students majoring in airline services improve TOEIC scores significantly (Kim et al., 2022). Employing a two-way Analysis of Variance (ANOVA), the research used a Quasi-experiment to find out how the AI-based program helped 119 participants improve TOEIC test scores. Researchers found that airline service students can increase their TOEIC scores through AI-assisted TOEIC learning programs (Kim et al., 2022). While using pre-and post-test designs to assess progress in TOEIC grammar and vocabulary, the other study also investigated that 121 students using an AI-based program to get better TOEIC test scores for a semester improved their vocabulary and understanding of English grammar (Kim et al., 2022). However, these studies had some limitations like small sample sizes and short-term effects. Future research suggests the need for a bigger sample size and a long-term study to generalize its key findings. This research lets us understand how personalized AI tutors improve students' language abilities.

## 2.4. AI Tutors and Students' Satisfaction

The increasing integration of Artificial Intelligence (AI) into language learning highlights the necessity to address user satisfaction, as it represents how students regulate learning through AI (Xia et al., 2023). To assess this, Yildiz (2023) developed a comprehensive scale to measure satisfaction levels, focusing on the AI's communicative, behavioral, and cognitive skills development. Based on the pilot survey conducted with 30 individuals learning the Turkish language, the study involved approximately 174 participants selected by the university to assess the time they spent on individual language learning programs such as Duolingo and ChatGPT. Results indicated that many individuals spend at least 1-3 hours daily on language learning programs. Higher user satisfaction was associated with the AI's communicative, behavioral, and cognitive elements during the language learning process.

Additionally, Das et al. (2023) analyzed that the satisfactory point of AI tutors lies in their ability to respond quickly with appropriate answers, helping users feel more in control of their learning pace. In such a way, users are further motivated and invested in their learning efforts, enriching their learning experience. Learning with AIs has garnered favorable satisfaction among users even in other subjects besides language learning (Wang et al., 2021). Moral-Sánchez et al. (2023) investigated student satisfaction by integrating chatbots into social networks they use for math education. Students were then assigned the task of creating their chatbots related to the subject and were asked to share them with the rest of the class. Through this process, other students had the opportunity to try out the new AI and provide feedback based on its performance. The responses were subsequently analyzed to extract concrete data, leading to the inference that many students were satisfied with the generated bots as educational aids, with satisfaction ratings falling in the range of 4 to 5 out of 5. Additionally, over 80% of the students reported that the activity of creating and sharing chatbots helped them become more familiar with digital technology, expressing a willingness to engage in similar activities with subjects beyond mathematics.

There was also an experiment conducted with a virtual tutoring system, as reported by Afzal et al. (2019). The team had an IBM Watson Tutor engaging with students and collecting their impressions about the AI tutor. Among 29 students, 78% of them felt that the AI tutor was relatively sincere in its engagement, and 60% of them agreed that the AI tutor certainly helped their personalized learning experiences. Furthermore, while not strictly rooted in Artificial Intelligence itself, virtual education making use of the metaverse has been reported to be satisfactory among students, as Hwang's (2022) report indicates. With 27 students making up 84% of all university student volunteers, they replied that they were mildly or greatly satisfied. Most of the students credited their satisfaction with factors like being easier to grasp the learning contents and indicating a sense of achievement. These findings underscore the significant potential of AI for enhancing learning engagement and satisfaction. Yildiz's (2023) finding, encompassing the positive aspects of communicative, behavioral, and cognitive, emerges as a valuable tool for assessing user experiences with language learning AI. Nevertheless, Yildiz emphasized the need to refine the satisfaction scale and explore its applicability across personalized language learning contexts and AI tools for further study. Understanding user preferences for specific AI features can guide the development of more effective language-learning tools, as high satisfaction of being in control of the learning pace leads to more motivation and engagement (Das et al., 2023). This was also supported by Hwang's research (2022) and research conducted with IBM Watson Tutor (Afzal et al., 2019), where many users attributed higher satisfaction with being easier to engage and helping them perform. Thus, the positive satisfaction rate of AI in personalized language learning underscores its promising potential to revolutionize the learning landscape.

## 3. Research Question

### 3.1. Research Purpose

This study aims to investigate how personalized AI-Tutors within language learning platforms, such as Santa and Duolingo, influence students' engagement, academic performance, and students' satisfaction in the learning material.

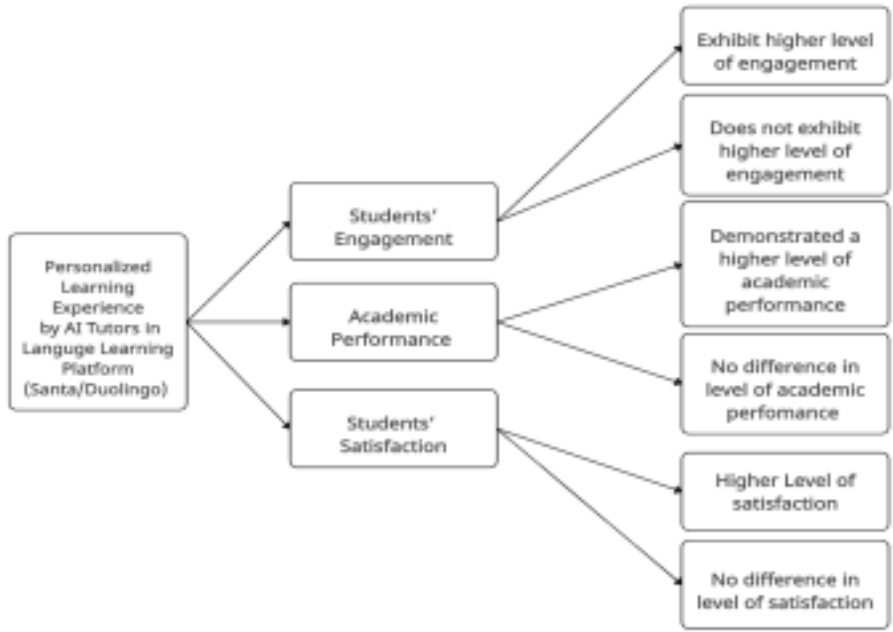

Figure 2: Personalized Language Learning Model

### 3.2. Research Question

How does personalized language learning experience facilitated by AI tutors in language learning platforms such as Santa and Duolingo influence students' engagement, academic performance, and students' satisfaction in the learning material?

- X (Intervention) - Students' personalized learning experience facilitated by AI tutors in language learning platforms such as Santa and Duolingo
- Y (Variable) - Students' Engagement, Academic Performance, and Students' Satisfaction in the learning material

|  | How does personalized learning experience facilitated by AI tutors in language learning platforms like Santa and Duolingo influence students' engagement, academic performance, and students' satisfaction in the learning material? |
|---|---|
| Intervention (X) | Students' personalized learning experience facilitated by AI tutors in language learning platforms like Santa and Duolingo |
| Variable (Y1) | Difference in Students' Engagement: Measured in a Scale of 0-10 0 (Not Engaging) - 10 (Very Engaging) |

| Variable (Y2) | Academic Performance:<br>Measured in five discrete grades that will be converted to numeric format (A : 5 / B : 4 / C : 3 / D : 2 / F : 1) |
|---|---|
| Variable (Y3) | Students' Satisfaction:<br>Measured in a Scale of 0-10<br>0 (Not Satisfied) - 10 (Very Satisfied) |

Table 1: Operational Definition of Variables

### 3.3. Hypothesis

$H_1$: There is a difference in the level of engagement with the learning material before and after the use of AI tutors.

$H_2$: There is a difference in the academic performance before and after the use of AI tutors.

$H_3$: There is a difference in the level of satisfaction with the learning material before and after the use of AI tutors.

To test our hypothesis, we will conduct a quasi-experiment utilizing a paired t-test with a pre-post test design. Our study will involve 34 new users of language learning platforms, and we will employ statistical analyses to compare the outcomes before and after the use of AI tutors in language learning platforms such as Santa and Duolingo. The results will allow us to examine the differences in students' engagement, academic performance, and learning satisfaction during a personalized language learning experience facilitated by AI tutors.

## 4. Methodology

### 4.1. Research Design

For this experiment, we selected Duolingo and Santa as the AI language learning platforms. Duolingo is considered one of the best AI tutor applications for improving speaking, listening, and vocabulary skills in the US market (Handini, 2022). As of 2023, Duolingo has reported 24.2 million daily active users and 83.1 million active users (Duolingo, 2023). On the other hand, Santa is one of the most popular AI tutoring platforms in Korea, designed to help users enhance their TOEIC scores and English proficiency (Kim et al., 2022). As of 2023, Santa has reported having a total of more than 5 million downloads (Kim et al., 2023). By opting for these two popular platforms created in different countries, we anticipated that we could easily find a diverse sampling, while ensuring a balanced and unbiased representation of demographics within our sample.

This study employs a pre-test/post-test quasi-experiment with a paired t-test design to evaluate the effectiveness of an AI tutor. Participants will complete a pre-test measuring their engagement level, academic performance, and satisfaction level before using AI tutor platforms like Santa and Duolingo. Afterward, they will complete a post-test measuring the same variable again. This one group pretest - posttest design (shown in Figure 3) (Creswel et al., 2018) allows us to assess individual change within the same group, minimizing variability due to pre-existing differences between participants. A paired t-test will be used to statistically analyze the paired pre-test and post-test data, revealing any significant changes in the variable associated with after using the AI tutor.

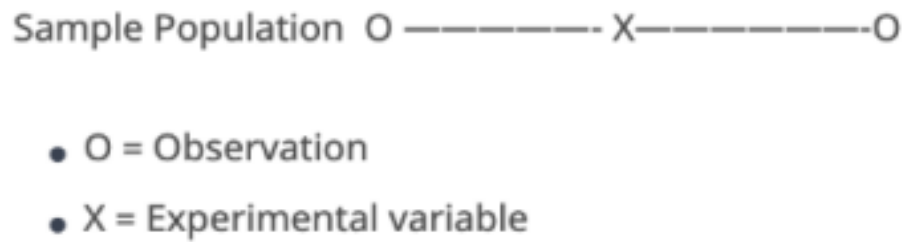

- O = Observation
- X = Experimental variable

Figure 3 : Quasi-Experimental Design - One Group Pretest - Posttest Design (Creswell et al., 2018, pg.269)

## 4.2. Study Population and Sampling

The participants for this experiment will include 34 students who are new users of AI language learning platforms like Santa and Duolingo. To determine the appropriate sample size, a power analysis was conducted utilizing a Sample Size Calculator (shown in Figure 4) (Ristl, 2024). The analysis considered the following variables: Mean Difference: 1, Standard Deviation: 2, $\alpha$ : 0.05, and Power: 0.8. This power analysis ensured that the study had adequate statistical power to detect a meaningful relationship if it exists. With these parameters, the analysis indicated that a minimum sample size of 34 students is needed for the paired t-test.

Figure 4: Power Analysis Variables showing the sample size for the paired-sample t test. (Ristl et al., 2024)

### 4.3. Data Collection

We will employ a web-based questionnaire (See Appendix A) to collect data. Employing a cross-sectional survey approach, we aim to capture a comprehensive picture of participants' experiences with AI tutoring platforms. The questionnaire will be structured into two distinct segments: a pre-use section and a post-use section. The pre-use segment is designed to establish a baseline, gathering information on participants' engagement, academic performance, and satisfaction before their interaction with AI platforms. Engagement will be measured by asking participants to rate their current engagement level with language learning on a scale of 0 - 10. To assess academic performance, participants will be prompted to provide measurable indicators such as test scores. Satisfaction will be measured by asking questions about participants' current satisfaction with language learning on a scale of 0-10.

In the post-use segment, participants will respond to the same set of questions, reflecting their academic performance, engagement, and satisfaction levels after their interaction with AI platforms. This structured approach will allow us to analyze changes and trends, providing valuable insights into the impact of AI tutors on language learning experiences.

### 4.4. Data Procedure

Following the data collection, we will calculate the mean difference between pre- and post-tests for each participant. To quantify changes in engagement and satisfaction levels, we will subtract participants' post-test scores from their pre-test scores, both of which were measured on a scale of 0-10. For academic performance, we will assign numerical values to corresponding letter grades: A (5), B (4), C (3), D (2), and F (1). The pre-test and post-test scores will be converted into these numerical values based on the participants' letter grades. The mean difference for each participant will be computed by subtracting the pre-test score from the post-test score for engagement, satisfaction, and academic performance. This approach will allow us to comprehensively analyze the impact of AI tutors on participants' language learning experiences.

## 4.5. Data Analysis

The primary statistical instrument employed in this study is the two tailed paired t-test. This test was specifically chosen to analyze the mean differences between pre- and post-tests for each variable, providing a quantitative measure of the impact of AI tutors on language learning experiences. For the hypotheses, a mean difference greater than zero for each variable would provide evidence supporting the idea that AI tutors have an impact on different levels of engagement, academic performance, and satisfaction in language learning. Conversely, if the mean difference for each variable equals zero, we will accept the null hypothesis. It is essential to acknowledge that the actual analysis for this experiment may yield more intricate and definitive outcomes. Future inferential analyses could investigate the intricate relationships among the three variables. Nevertheless, drawing on insights from the literature review, it is reasonable to hypothesize that a power analysis indicates support for the alternative hypotheses across all three variables: students' engagement, academic performance, and students' satisfaction. This validates that AI-driven personalized learning on platforms like Santa and Duolingo significantly enhances students' engagement, academic performance, and satisfaction.

## 5. Discussion

### 5.1 Research Limitations

This study is based on a hypothetical research proposal, and as such, the methodology and data collection processes were not implemented in practice. Consequently, the research lacks empirical evidence and real-world data to support its assumptions, analysis, or projected outcomes. The study's findings and implications are therefore speculative and serve primarily as a conceptual framework for future investigation.

In future research, it will be essential to conduct actual data collection through the proposed testing above to validate the proposed hypotheses. This will also allow for a more rigorous evaluation of the research questions and provide empirical support for the conclusions. Furthermore, any practical or logistical challenges associated with data gathering, participant recruitment, or analytical limitations can only be properly addressed in a real-world research context.

## 6. Conclusion

This study examined the influence of personalized AI tutors on language learners' engagement, academic performance, and satisfaction. The purpose of our research was to provide valuable insights for researchers and educators in the field of language learning. Our findings resonated with existing research that highlights the potential of AI to enhance engagement, academic performance, and satisfaction in the learning process (Kim et al., 2020;

Yildiz, 2023; Moral-Sánchez et al., 2023). The existing research indicates that the AI tutor providing personalized curriculum and systematic academic analysis to students significantly increases users' engagement levels, academic performance, and satisfaction. Our research question: How does personalized language learning experience facilitated by AI tutors in language learning platforms influence student's engagement, academic performance, and student's satisfaction in the learning material is directly related to the finding in the earlier research. In previous research, a small sample size and short-term study may not have been enough for generalization. To address this research gap, we will conduct multiple surveys to capture a bigger sample size related to users' experience with AI tutoring platforms before and after. This research significantly addressed the gap by exploring the specific impact of AI tutors within the context of language learning. Our findings provide concrete evidence for the positive influence of AI tutors, showcasing their ability to elevate engagement, performance, and satisfaction. This aligns with our initial hypotheses and contributes to the understanding of AI's role in education. Looking forward, examining the long-term effects of AI on academic achievement and learner motivation remains crucial. Additionally, understanding user preferences for specific AI features can guide the development of even more effective language-learning tools. While limitations such as specific age groups and small sample size restrict the generalizability of our findings to other areas of study, future research expanding participant pools can build upon these initial insights.

# Appendix A

https://forms.gle/Mi3zdtmSmyf4mu9P6

# References

Afzal, S., Dempsey, B., D'Helon, C., Mukhi, N., Pribic, M., Sickler, A., Strong, P., Vanchiswar, M., & Wilde, L. (2019). The Personality of AI Systems in Education: Experiences with the Watson Tutor, a one-on-one virtual tutoring system. *The Personality of AI Systems in Education: Experiences With the Watson Tutor, a One-on-one Virtual Tutoring System*, 95(1), 44–52. https://doi.org/10.1080/00094056.2019.1565809

Ballan, D. (2023, December 11). *Revolutionizing education: The evolution of modern teaching techniques*. English Plus Podcast. https://englishpluspodcast.com/revolutionizing-education-the-evolution-of-modern-teaching-techniques/

Beck, J., Stern, M., & Haugsjaa, E. (1996). Applications of AI in education. *XRDS Crossroads The ACM Magazine for Students*, *3*(1), 11–15. https://doi.org/10.1145/332148.332153

Creswell, J. W., & Creswell, J. D. (2018). *Research design* (5th ed.). SAGE Publications.

Das, A., Malaviya, S., & Singh, M. (2023, August). International Journal of Computer Sciences and Engineering. ijcseonline.org. https://doi.org/10.26438/ijcse/v11i8.1522

Davis, K., Seider, S., & Christodoulou, J. (2011). *(PDF) The theory of multiple intelligences - researchgate*. The Theory of Multiple Intelligences. https://www.researchgate.net/publication/317388610_The_Theory_of_Multiple_Intelligences

Diwan, C., Srinivasa, S., Suri, G., Agarwal, S., & Ram, P. (2022, December 7). *AI-Based Learning Content Generation and learning pathway augmentation to increase learner engagement*. Computers and Education: Artificial Intelligence. https://doi.org/10.1016/j.caeai.2022.100110

Duolingo. (2023). *Q3fy23 shareholder letter - investors.duolingo.com*. shareholder letter Q3 2023. https://investors.duolingo.com/static-files/033ccf0a-f5ea-4897-ba7a-00b00bb48a2d

Handini, B. S. (2022). The effect of artificial intelligent technology used (duolingo ... The Effect of Artificial Intelligent Technology Used (Duolingo Application) To Enhance English Learning. http://doi.org/10.32528/ellite.v7i2.8354

Huang, A. Y. Q., Lu, O. H. T., & Yang, S. J. H. (2022, November 26). Effects of artificial intelligence–enabled personalized recommendations on learners' learning engagement, motivation, and outcomes in a flipped classroom. Effects of artificial Intelligence–Enabled personalized recommendations on learners' learning engagement, motivation, and outcomes in a flipped classroom. https://doi.org/10.1016/j.compedu.2022.104684

Kim, B., Suh, H., Heo, J., & Choi, Y. (2020). AI-driven interface design for Intelligent Tutoring System improves student engagement. In *arXiv [cs.CY]*. https://doi.org/10.48550/arXiv.2009.08976

Kim, H. S., Kim, N. Y., & Cha, Y. (2022). A Study on the Use of AI-based Learning Programs by EFL Students with Different Types of Teacher Support. 22, 355-376. https://doi.org/10.15738/kjell.22..202204.355

Kim, J.-W. (2023, June 2). *Riiid inks deal with ALC to expand its AI-based TOEIC platform in Japan*. KED Global. https://www.kedglobal.com/artificial-intelligence/newsView/ked202306020002

Kim, N. Y. (2022). English with AI: A new era of TOEIC learning for students majoring in airline services. Linguistic Research, 39(3). https://doi.org/10.17250/khisli.39..202209.004

Kim, W.-H., & Kim, J.-H. (2020). *Individualized AI Tutor Based on Developmental Learning Networks*. Ieee.org. https://doi.org/10.1109/ACCESS.2020.2972167

Kokku, R., Sundararajan, S., Dey, P., Sindhgatta, R., Nitta, S., & Sengupta, B. (2018). *Augmenting Classrooms with AI for Personalized Education*. Ieee.org. https://doi.org/10.1109/ICASSP.2018.8461812

LEARNING APP TO IMPROVE L2 LEARNERS' FLUENCY". Masters Theses. 120. https://doi.org/10.7275/28641843

Lopez, D. M., & Schroeder, L. (2008, May). DESIGNING STRATEGIES THAT MEET THE VARIETY OF LEARNING STYLES OF STUDENTS . https://eric.ed.gov/?id=ED500848

Minn, S., a, & b. (2022, February 7). *Ai-Assisted Knowledge Assessment Techniques for Adaptive Learning Environments*. Computers and Education: Artificial Intelligence. https://doi.org/10.1016/j.caeai.2022.100050 Received 27 July 2021; Received in revised form

Mollick, E., & Mollick, L. (2023, June 11). *Assigning AI: Seven approaches for students with prompts - arxiv.org*. ASSIGNING AI: SEVEN APPROACHES FOR STUDENTS WITH PROMPTS. https://doi.org/10.48550/arXiv.2306.10052

Moral-Sánchez, S. N., Rey, F. J. R., & Cebrián-De-La-Serna, M. (2023). Analysis of artificial intelligence chatbots and satisfaction for learning in mathematics education. *ResearchGate*. https://doi.org/10.46661/ijeri.8196

Morgan, H. (2021). *Howard Gardner's multiple intelligences theory and his ideas on ... - ed*. HOWARD GARDNER'S MULTIPLE INTELLIGENCES THEORY AND HIS IDEAS ON PROMOTING CREATIVITY HANI MORGAN. https://eric.ed.gov/?id=ED618540

Nakayama, Ryo (2022) "TALKING TO AI TUTORS: SPEAKING PRACTICE USING A JAPANESE LANGUAGE. https://doi.org/10.7275/28641843

Rane, N., Choudhary, S., & Rane, J. (2023, November 29). *Education 4.0 and 5.0: Integrating Artificial Intelligence (AI) for personalized and adaptive learning*. SSRN. http://dx.doi.org/10.2139/ssrn.4638365

Ristl, R. (2024). Sample size calculator. https://homepage.univie.ac.at/robin.ristl/samplesize.php?test=pairedttest

Ruan, S., Davis, G. M., Jiang, L., Liu, Z., Landay, J. A., Xu, Q., & Brunskill, E. (2021). *Stanford HCI Group*. EnglishBot: An AI-Powered Conversational System for Second Language Learning. https://doi.org/10.1145/3397481.3450648

Sedlmeier , P. (2002, November 2). *Intelligent Tutoring Systems*. International Encyclopedia of the Social & Behavioral Sciences. https://doi.org/10.1016/B0-08-043076-7/01618-1

Tetzlaff, L., Schmiedek, F., & Brod, G. (2020, October 29). *Developing personalized education: A Dynamic Framework - Educational Psychology Review*. SpringerLink. https://doi.org/10.1007/s10648-020-09570-w

Thomas K.F. Chiu, Qi Xia, Xinyan Zhou, Ching Sing Chai, Miaoting Cheng. (2023). *Systematic literature review on opportunities, challenges, and future research recommendations of artificial intelligence in education*. Sciencedirect.com. https://doi.org/10.1016/j.caeai.2022.100118

Vygotsky, L. S. (1930). Mind in Society: The Development of Higher Psychological Processes. Cambridge, MA: Harvard University Press. https://doi.org/10.2307/j.ctvjf9vz4

Wang, H., Tlili, A., Huang, R. et al. (2021) Examining the applications of intelligent tutoring

systems in real educational contexts: A systematic literature review from the social experiment perspective. Educ Inf Technol 28, 9113–9148. https://doi.org/10.1007/s10639-022-11555-x

Woo, J. H., & Choi, H. (2021). Systematic review for AI-based language learning tools. *Journal of Digital Contents Society*, *22*(11), 1783–1792. https://doi.org/10.9728/dcs.2021.22.11.1783

Xia, Q., Chiu, T. K. F., Chai, C. S., & Xie, K. (2023). The mediating effects of needs satisfaction on the relationships between prior knowledge and self-regulated learning through artificial intelligence chatbot. https://doi.org/10.1111/bjet.13305

Xu, X., Dugdale, D. M., Wei, X., & Mi, W. (2022). *Full article: Leveraging Artificial Intelligence to predict young ...* Leveraging Artificial Intelligence to Predict Young Learner Online Learning Engagement. https://doi.org/10.1080/08923647.2022.2044663

Yıldız, T. (2023). Measurement of Attitude in Language Learning with AI (MALL:AI). *Participatory Educational Research*, *10*(4), 111–126. https://doi.org/10.17275/per.23.62.10.4